\documentclass[10pt,twocolumn,letterpaper]{article}

\usepackage{tikz}
\usepackage{iccv}
\usepackage{times}
\usepackage{epsfig}
\usepackage{graphicx}
\usepackage{amsmath}
\usepackage{amssymb}
\usepackage{booktabs}

\usepackage{balance}
\usepackage{microtype}
\usepackage{xcolor}
\usepackage[ruled,vlined]{algorithm2e}
\usepackage{pifont}
\usetikzlibrary{fit,calc}
\newcommand*{\tikzmk}[1]{\tikz[remember picture,overlay,] \node (#1) {};\ignorespaces}
\newcommand{\boxit}[1]{\tikz[remember picture,overlay]{\node[yshift=3pt,fill=#1,opacity=.25,fit={(A)($(B)+(.71\linewidth,.8\baselineskip)$)}] {};}\ignorespaces}
\newcommand{\boxits}[1]{\tikz[remember picture,overlay]{\node[yshift=3pt,fill=#1,opacity=.25,fit={(A)($(B)+(.64\linewidth,.8\baselineskip)$)}] {};}\ignorespaces}
\colorlet{pink}{red!40}
\colorlet{blue}{cyan!60}
\colorlet{greenl}{green!40}

\usepackage[pagebackref=true,breaklinks=true,letterpaper=true,colorlinks,bookmarks=false]{hyperref}

\iccvfinalcopy % *** Uncomment this line for the final submission

\def\iccvPaperID{11265} % *** Enter the ICCV Paper ID here

\ificcvfinal\fi

\definecolor{red}{rgb}{.73, 0.12, 0.14}
\definecolor{blue}{rgb}{.2, 0.45, 0.7}
\definecolor{orange}{rgb}{.70, 0.3, 0.15}
\definecolor{green}{rgb}{.553, 0.741, 0.522}

\newcommand{\ours}[0]{DietNeRF}

\newcommand{\ourspixel}[0]{DietPixelNeRF}
\newcommand{\Lmse}[0]{\mathcal{L}_\text{MSE}}
\newcommand{\Lsc}[0]{\mathcal{L}_\text{SC}}
\newcommand{\cmark}{\ding{51}}%
\newcommand{\xmark}{\ding{55}}%

\begin{document}

%%%%%%%%% TITLE
\title{Putting NeRF on a Diet: Semantically Consistent Few-Shot View Synthesis}

\author{Ajay Jain\\
UC Berkeley\\
{\tt\small ajayj@berkeley.edu}
\and
Matthew Tancik\\
UC Berkeley\\
{\tt\small tancik@berkeley.edu}
\and
Pieter Abbeel\\
UC Berkeley\\
{\tt\small pabbeel@cs.berkeley.edu}
}

\maketitle
% Remove page # from the first page of camera-ready.
\ificcvfinal\thispagestyle{empty}\fi

%%%%%%%%% ABSTRACT
\begin{abstract}
We present \ours{}, a 3D neural scene representation estimated from a few images.
Neural Radiance Fields (NeRF) learn a continuous volumetric representation of a scene through multi-view consistency, and can be rendered from novel viewpoints by ray casting. While NeRF has an impressive ability to reconstruct geometry and fine details given many images, up to 100 for challenging 360$^\circ$ scenes, it often finds a degenerate solution to its image reconstruction objective when only a few input views are available. To improve few-shot quality, we propose \ours{}. We introduce an auxiliary semantic consistency loss that encourages realistic renderings at novel poses. \ours{} is trained on individual scenes to (1) correctly render given input views from the same pose, and (2) match high-level semantic attributes across different, random poses. Our semantic loss allows us to supervise \ours{} from arbitrary poses. We extract these semantics using a pre-trained visual encoder such as CLIP, a Vision Transformer trained on hundreds of millions of diverse single-view, 2D photographs mined from the web with natural language supervision.
In experiments, \ours{} improves the perceptual quality of few-shot view synthesis when learned from scratch, can render novel views with as few as one observed image when pre-trained on a multi-view dataset, and produces plausible completions of completely unobserved regions.
Our project website is available at {\small\url{https://www.ajayj.com/dietnerf}}.
\end{abstract}

%%%%%%%%% BODY TEXT
\section{Introduction}

\begin{figure}
    \centering
    \includegraphics[width=\columnwidth]{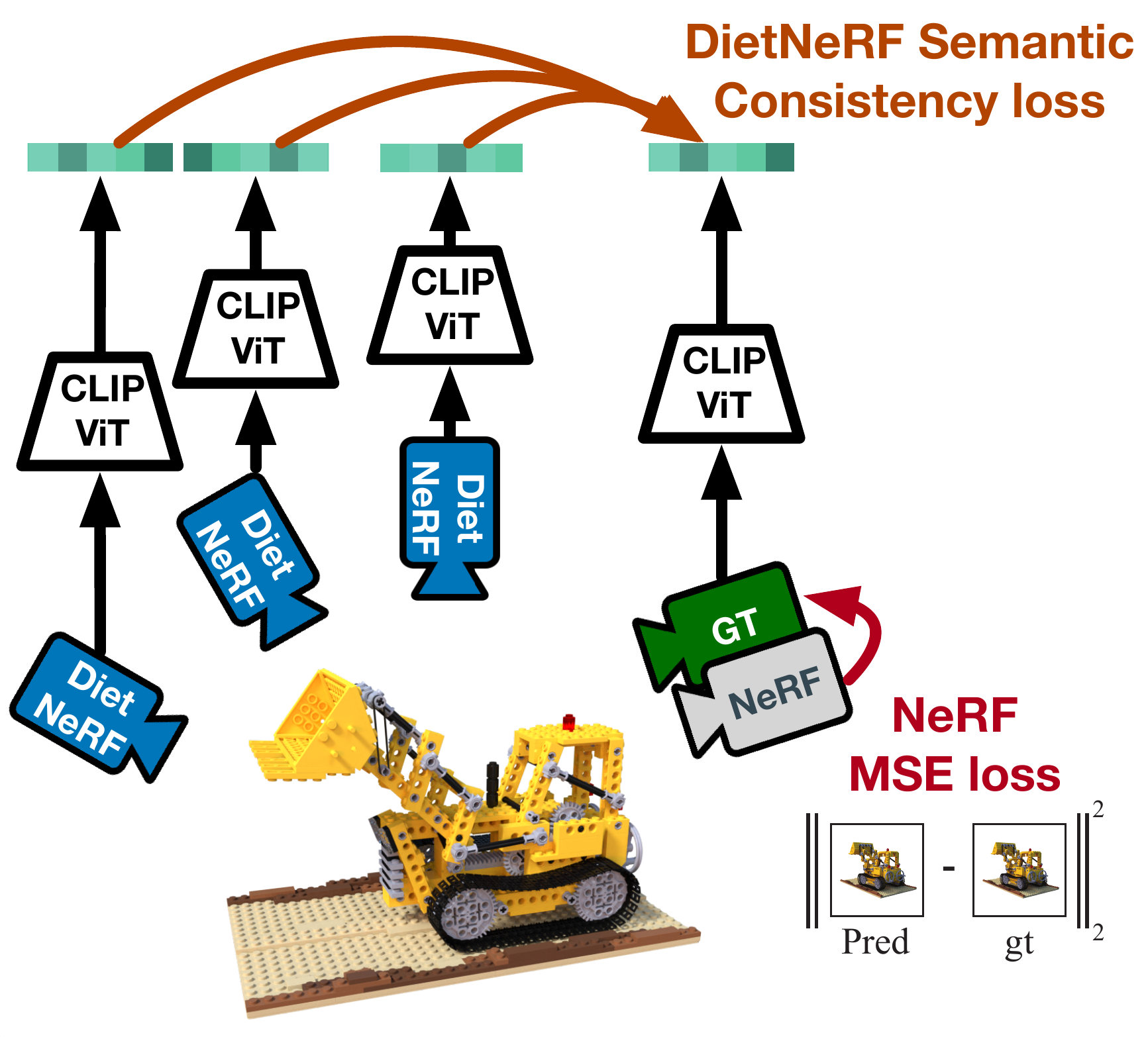}
    \caption{Neural Radiance Fields are trained to represent a scene by supervising renderings from the \textit{same pose} as ground-truth observations (\textbf{\color{red} MSE loss}). However, when only a few views are available, the problem is underconstrained. NeRF often finds degenerate solutions unless heavily regularized. Based on the principle that
    \textbf{``a bulldozer is a bulldozer from any perspective''},
    our proposed \ours{} supervises the radiance field from arbitrary poses (\textbf{\color{blue} \ours{} cameras}). This is possible because we compute a \textbf{\color{orange} semantic consistency loss} in a feature space capturing high-level scene attributes, not in pixel space. We extract semantic representations of renderings using the CLIP Vision Transformer~\cite{radford2021learning}, then maximize similarity with representations of ground-truth views. In effect, we use prior knowledge about scene semantics learned by \textit{single-view} 2D image encoders to constrain a 3D representation.}
    \label{fig:teaser}
\end{figure}

\begin{figure*}[t]
    \centering
    \includegraphics[width=\linewidth]{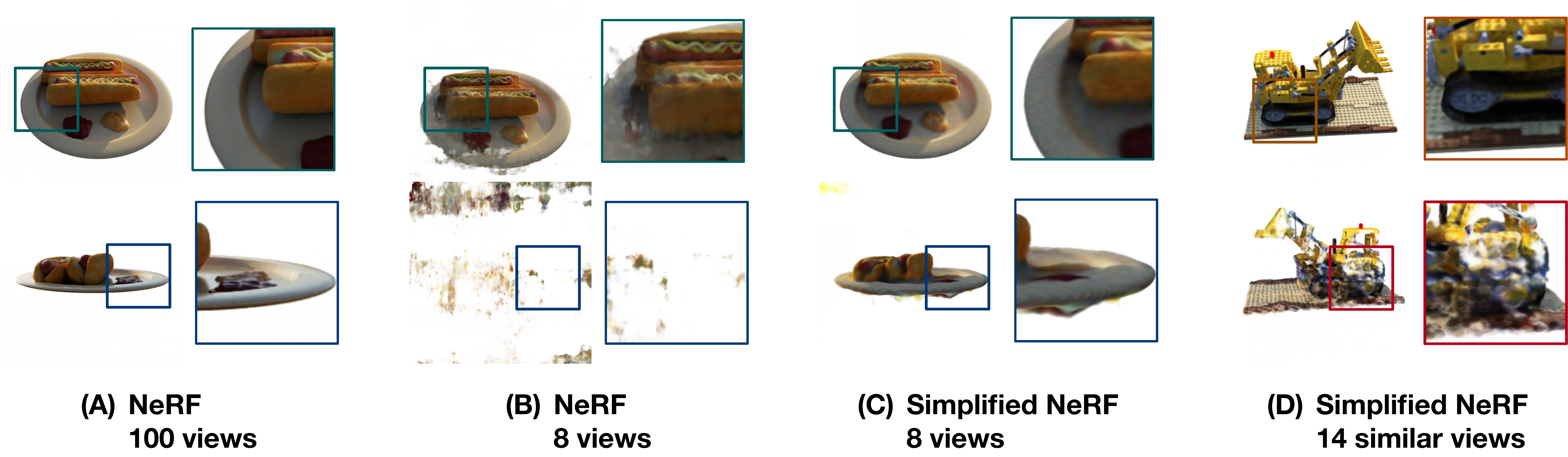}
    \caption{\textbf{Few-shot view synthesis is a challenging problem for Neural Radiance Fields.} \textbf{(A)} When we have 100 observations of an object from uniformly sampled poses, NeRF estimates a detailed and accurate representation that allows for high-quality view synthesis purely from multi-view consistency. \textbf{(B)} However, with only 8 views, the same NeRF overfits by placing the object in the near-field of the training cameras, leading to misplaced objects at poses near training cameras and degeneracies at novel poses. \textbf{(C)} We find that NeRF can converge when regularized, simplified, tuned and manually reinitialized, but no longer captures fine details. \textbf{(D)} Finally, without prior knowledge about similar objects, single-scene view synthesis cannot plausibly complete unobserved regions, such as the left side of an object seen from the right. In this work, we find that \textbf{these failures occur because NeRF is only supervised from the sparse training poses.}}
    \label{fig:challenges}
\end{figure*}

In the novel view synthesis problem, we seek to rerender a scene from arbitrary viewpoint given a set of sparsely sampled viewpoints. View synthesis is a challenging problem that requires some degree of 3D reconstruction in addition to high-frequency texture synthesis.
Recently, great progress has been made on high-quality view synthesis when many observations are available. A popular approach is to use Neural Radiance Fields (NeRF)~\cite{mildenhall2020nerf} to estimate a continuous neural scene representation from image observations. During training on a particular scene, the representation is rendered from observed viewpoints using volumetric ray casting to compute a reconstruction loss. At test time, NeRF can be rendered from novel viewpoints by the same procedure. While conceptually very simple, NeRF can learn high-frequency view-dependent scene appearances and accurate geometries that allow for high-quality rendering.

Still, NeRF is estimated per-scene, and cannot benefit from prior knowledge acquired from other images and objects. Because of the lack of prior knowledge, NeRF requires a large number of input views to reconstruct a given scene at high-quality. 
Given 8 views, Figure~\ref{fig:challenges}B shows that novel views rendered with the full NeRF model contain many artifacts because the optimization finds a degenerate solution that is only accurate at observed poses.
We find that the core issue is that prior 3D reconstruction systems based on rendering losses are \textit{only supervised at known poses}, so they overfit when few poses are observed. Regularizing NeRF by simplifying the architecture avoids the worst artifacts, but comes at the cost of fine-grained detail.

Further, prior knowledge is needed when the scene reconstruction problem is underdetermined. 3D reconstruction systems struggle when regions of an object are never observed. This is particularly problematic when rendering an object at significantly different poses. When rendering a scene with an extreme baseline change, unobserved regions during training become visible. A view synthesis system should generate plausible missing details to fill in the gaps.
Even a regularized NeRF learns poor extrapolations to unseen regions due to its lack of prior knowledge (Figure~\ref{fig:challenges}D).

Recent work trained NeRF on \textit{multi-view} datasets of similar scenes~\cite{yu2020pixelnerf, grf2020, schwarz2020graf, tancik2020learned, wang2021ibrnet} to bias reconstructions of novel scenes. Unfortunately, these models often produce blurry images due to uncertainty, or are restricted to a single object category such as ShapeNet classes as it is challenging to capture large, diverse, multi-view data.

In this work, we exploit the consistency principle that \textit{``a bulldozer is a bulldozer from any perspective''}: objects share high-level semantic properties between their views. 
Image recognition models learn to extract many such high-level semantic features including object identity. We transfer prior knowledge from pre-trained image encoders learned on highly diverse 2D \textit{single-view} image data to the view synthesis problem. In the single-view setting, such encoders are frequently trained on millions of realistic images like ImageNet~\cite{deng2009imagenet}. CLIP is a recent multi-modal encoder that is trained to match images with captions in a massive web scrape containing 400M images~\cite{radford2021learning}. Due to the diversity of its data, CLIP showed promising zero- and few-shot transfer performance to image recognition tasks. We find that CLIP and ImageNet models also contain prior knowledge useful for novel view synthesis.

We propose \ours{}, a neural scene representation based on NeRF that can be estimated from only a few photos, and can generate views with unobserved regions. In addition to minimizing NeRF's mean squared error losses at known poses in pixel-space, \ours{} penalizes a \textit{semantic consistency} loss. This loss matches the final activations of CLIP's Vision Transformer~\cite{dosovitskiy2021an} between ground-truth images and rendered images at \textit{different} poses, allowing us to supervise the radiance field from arbitrary poses. 
In experiments, we show that \ours{} learns realistic reconstructions of objects with as few as 8 views without simplifying the underlying volumetric representation, and can even produce reasonable reconstructions of completely occluded regions. To generate novel views with as few as 1 observation, we fine-tune pixelNeRF~\cite{yu2020pixelnerf}, a generalizable scene representation, and improve perceptual quality.

\section{Background on Neural Radiance Fields}

A plenoptic function, or light field, is a five-dimensional function that describes the light radiating from every point in every direction in a volume such as a bounded scene. While explicitly storing or estimating the plenoptic function at high resolution is impractical due to the dimensionality of the input, Neural Radiance Fields~\cite{mildenhall2020nerf} parameterize the function with a continuous neural network such as a multi-layer perceptron (MLP). A Neural Radiance Field (NeRF) model is a five-dimensional function $f_\theta(\mathbf{x}, \mathbf{d}) = (\mathbf{c}, \sigma)$ of spatial position $\mathbf{x}=(x,y,z)$ and viewing direction $(\theta, \phi)$, expressed as a 3D unit vector $\mathbf{d}$. NeRF predicts the RGB color $\mathbf{c}$ and differential volume density $\sigma$ from these inputs. To encourage view-consistency, the volume density only depends on $\mathbf{x}$, while the color also depends on viewing direction $\mathbf{d}$ to capture viewpoint dependent effects like specular reflections.
Images are rendered from a virtual camera at any position by integrating color along rays cast from the observer according to volume rendering~\cite{10.1145/800031.808594}:
\begin{equation}
    \mathbf{C}(\mathbf{r}) = \int_{t_n}^{t_f} T(t)\sigma(\mathbf{r}(t))\mathbf{c}(\mathbf{r}(t), \mathbf{d})dt
    \label{eq:volume_integration}
\end{equation}
where the ray originating at the camera origin $\mathbf{o}$ follows path $\mathbf{r}(t) = \mathbf{o} + t\mathbf{d}$, and the transmittance ${T(t)=\exp \left(- \int_{t_n}^{t_f} \sigma(\mathbf{r}(s)) ds\right)}$ weights the radiance by the probability that the ray travels from the image plane at $t_n$ to $t$ unobstructed. To approximate the integral, NeRF employs a hierarchical sampling algorithm to select function evaluation points near object surfaces along each ray. NeRF separately estimates two MLPs, a coarse network and a fine network, and uses the coarse network to guide sampling along the ray for more accurately estimating~\eqref{eq:volume_integration}.
The networks are trained from scratch on \textit{each scene} given tens to hundreds of photos from various perspectives. Given observed multi-view training images $\{I_i\}$ of a scene, NeRF uses COLMAP SfM~\cite{Schonberger_2016_CVPR} to estimate camera extrinsics (rotations and origins) $\{\mathbf{p}_i\}$, creating a posed dataset $\mathcal{D}=\{(I_i, \mathbf{p}_i)\}$.

\section{NeRF Struggles at Few-Shot View Synthesis}

View synthesis is a challenging problem when a scene is only sparsely observed. Systems like NeRF that train on individual scenes especially struggle without prior knowledge acquired from similar scenes. We find that NeRF fails at few-shot novel view synthesis in several settings.

\textbf{NeRF overfits to training views}~~
Conceptually, NeRF is trained by mimicking the image-formation process at observed poses. The radiance field can be estimated repeatedly sampling a training image and pose $(I, \mathbf{p}_i)$, rendering an image $\hat{I}_{\mathbf{p}_i}$ from the \textbf{same pose} by volume integration \eqref{eq:volume_integration}, then minimizing the mean-squared error (MSE) between the images, which should align pixel-wise:
\begin{equation}
    \mathcal{L}_\text{full}(I, \hat{I}_{\mathbf{p}_i}) = \frac{1}{HW} \| I - \hat{I}_{\mathbf{p}_i} \|_2^2
\end{equation}
In practice, NeRF samples a smaller batch of rays across all training images to avoid the computational expense of rendering full images during training. Given subsampled rays $\mathcal{R}$ cast from the training cameras, NeRF minimizes:
\begin{equation}
    \mathcal{L}_\text{MSE}(\mathcal{R}) = \frac{1}{|\mathcal{R}|} \sum_{\mathbf{r} \in \mathcal{R}} \| \mathbf{C}(\mathbf{r}) - \hat{\mathbf{C}}(\mathbf{r}) \|_2^2
\end{equation}
With many training views, $\Lmse$ provides training signal to $f_\theta$ densely in the volume and does not overfit to individual training views. Instead, the MLP recovers accurate textures and occupancy that allow interpolations to new views (Figure~\ref{fig:challenges}A). Radiance fields with sinusoidal positional embeddings are quite effective at learning high-frequency functions~\cite{tancik2020learned}, which helps the MLP represent fine details.

Unfortunately, this high-frequency representational capacity allows NeRF to overfit to each input view when only a few are available. $\Lmse$ can be minimized by packing the reconstruction $\hat{I}_\mathbf{p}$ of training view $(I, \mathbf{p})$ close to the camera. Fundamentally, the plenoptic function representation suffers from a near-field ambiguity~\cite{kaizhang2020} where distant cameras each observe significant regions of space that no other camera observes. In this case, the optimal scene representation is underdetermined. Degenerate solutions can also exploit the view-dependence of the radiance field. Figure~\ref{fig:challenges}B shows novel views from the same NeRF trained on 8 views. While a rendered view from a pose near a training image has reasonable textures, it is skewed incorrectly and has cloudy artifacts from incorrect geometry. As the geometry is not estimated correctly, a distant view contains almost none of the correct information. High-opacity regions block the camera. Without supervision from any nearby camera, opacity is sensitive to random initialization.

\textbf{Regularization fixes geometry, but hurts fine-detail}~~
High-frequency artifacts such as spurious opacity and rapidly varying colors can be avoided in some cases by regularizing NeRF. We simplify the NeRF architecture by removing hierarchical sampling and learning only a single MLP, and reducing the maximum frequency positional embedding in the input layer. This biases NeRF toward lower frequency solutions, such as placing content in the center of the scene farther from the training cameras. We also can address some few-shot optimization challenges by lowering the learning rate to improve initial convergence, and manually restarting training if renderings are degenerate. Figure~\ref{fig:challenges}C shows that these regularizers successfully allow NeRF to recover plausible object geometry. However, high-frequency, fine details are lost compared to \ref{fig:challenges}A.

\textbf{No prior knowledge, no generalization to unseen views}~~
As NeRF is estimated from scratch per-scene, it has no prior knowledge about natural objects such as common symmetries and object parts. In Figure~\ref{fig:challenges}D, we show that NeRF trained with 14 views of the right half of a Lego vehicle generalizes poorly to its left side. We regularized NeRF to remove high-opacity regions that originally blocked the left side entirely. Even so, the essential challenge is that NeRF receives no supervisory signal from $\Lmse$ to the unobserved regions, and instead relies on the inductive bias of the MLP for any inpainting. We would like to introduce prior knowledge that allows NeRF to exploit bilateral symmetry for plausible completions.

\section{Semantically Consistent Radiance Fields}

Motivated by these challenges, we introduce the \ours{} scene representation. \ours{} uses prior knowledge from a pre-trained image encoder to guide the NeRF optimization process in the few-shot setting.

\newcommand{\Lscl}[0]{\mathcal{L}_{\text{SC}, \ell_2}}
\newcommand{\Lfull}[0]{\mathcal{L}_{\text{full}}}
\subsection{Semantic consistency loss}
\ours{} supervises $f_\theta$ at arbitrary camera poses during training with a semantic loss. While pixel-wise comparison between ground-truth observed images and rendered images with $\Lmse$ is only useful when the rendered image is aligned with the observed pose, humans are easily able to detect whether two images are views of the same object from semantic cues. We can in general compare a \textit{representation} of images captured from different viewpoints:
 \begin{equation}
     \Lscl(I, \hat{I}) = \frac{\lambda}{2} \|\phi(I) - \phi(\hat{I})\|_2^2
     \label{eq:Lscl}
 \end{equation}
If $\phi(x) = x$, Eq.~\eqref{eq:Lscl} reduces to $\Lfull$ up to a scaling factor. However, the identity mapping is view-dependent. We need a representation that is similar across views of the same object and captures important high-level semantic properties like object class.
We evaluate the utility of two sources of supervision for representation learning. First, we experiment with the recent CLIP model pre-trained for multi-modal language and vision reasoning with contrastive learning~\cite{radford2021learning}.
We then evaluate visual classifiers pre-trained on labeled ImageNet images~\cite{dosovitskiy2021an}. In both cases, we use similar Vision Transformer (ViT) architectures.

A Vision Transformer is appealing because its performance scales very well to large amounts of 2D data. 
Training on a large variety of images allows the network to encounter multiple views of an object class over the course of training without explicit multi-view data capture. It also allows us to transfer the visual encoder to diverse objects of interest in graphics applications, unlike prior class-specific reconstruction work that relies on homogeneous datasets~\cite{journals/pami/CashmanF13,cmrKanazawa18}.
ViT extracts features from non-overlapping image patches in its first layer, then aggregates increasingly abstract representations with Transformer blocks based on global self-attention~\cite{vaswani2017attention} to produce a single, global embedding vector. ViT outperformed CNN encoders in our early experiments.

In practice, CLIP produces normalized image embeddings. When $\phi(\cdot)$ is a unit vector, Eq.~\eqref{eq:Lscl} simplifies to cosine similarity up to a constant and a scaling factor that can be absorbed into the loss weight $\lambda$:
\begin{equation}
    \mathcal{L}_{\text{SC}}(I, \hat{I}) = \lambda \phi(I)^T \phi(\hat{I})
    \label{eq:Lsc}
\end{equation}
We refer to $\Lsc$~\eqref{eq:Lsc} as a \textit{semantic consistency} loss because it measures the similarity of high-level semantic features between observed and rendered views.
In principle, semantic consistency is a very general loss that can be applied to any 3D reconstruction system based on differentiable rendering.

\begin{algorithm}[t]
\footnotesize
\SetAlgoLined
\KwData{Observed views $\mathcal{D}=\{(I, \mathbf{p})\}$, semantic embedding function $\phi(\cdot)$, pose distribution $\pi$, consistency interval $K$,
weight $\lambda$, rendering size, batch size $|\mathcal{R}|$, lr $\eta_{it}$}
\KwResult{Trained Neural Radiance Field $f_\theta(\cdot, \cdot)$}
 Initialize NeRF $f_\theta(\cdot, \cdot)$\;
 \tikzmk{A}Pre-compute target embeddings $\{\phi(I) : I \in \mathcal{D}\}$\;\tikzmk{B}
  \boxit{greenl}
 \For{it from 1 to num\_iters}{
  Sample ray batch $\mathcal{R}$, ground-truth colors $\mathbf{C}(\cdot)$\;
  Render rays $\hat{\mathbf{C}}(\cdot)$ by~\eqref{eq:volume_integration}\;
  $\mathcal{L} \leftarrow \Lmse(\mathcal{R}, \mathbf{C}, \hat{\mathbf{C}})$\;
  \tikzmk{A}\If{$\text{it}~\%~K = 0$}{
  Sample target image, pose $(I, \mathbf{p})\sim\mathcal{D}$\;
  Sample source pose $\hat{\mathbf{p}} \sim \pi$\;
  Render image $\hat{I}$ from pose $\hat{\mathbf{p}}$\;
  $\mathcal{L} \leftarrow \mathcal{L} + \Lsc(I, \hat{I})$\;
  }\tikzmk{B}
  \boxits{greenl}
  Update parameters: $\theta \leftarrow Adam(\theta, \eta_{it}, \nabla_\theta \mathcal{L})$\;
 }
 \caption{Training \ours{} on a single scene}
 \label{alg:scarf}
\end{algorithm}

\subsection{Interpreting representations across views}
The pre-trained CLIP model that we use is trained on hundreds of millions of images with captions of varying detail. Image captions provide rich supervision for image representations. On one hand, short captions express semantically sparse learning signal as a flexible way to express labels~\cite{desai2021virtex}. For example, the caption ``A photo of hotdogs'' describes Fig.~\ref{fig:challenges}A. Language also provides semantically dense learning signal by describing object properties, relationships and appearances~\cite{desai2021virtex} such as the caption
``Two hotdogs on a plate with ketchup and mustard''. To be predictive of such captions, an image representation must capture some high-level semantics that are stable across viewpoints.
Concurrently, \cite{goh2021multimodal} found that CLIP representations capture visual attributes of images like art style and colors, as well as high-level semantic attributes including object tags and categories, facial expressions, typography, geography and brands.

\begin{figure}
    \centering
    \includegraphics[width=0.9\linewidth]{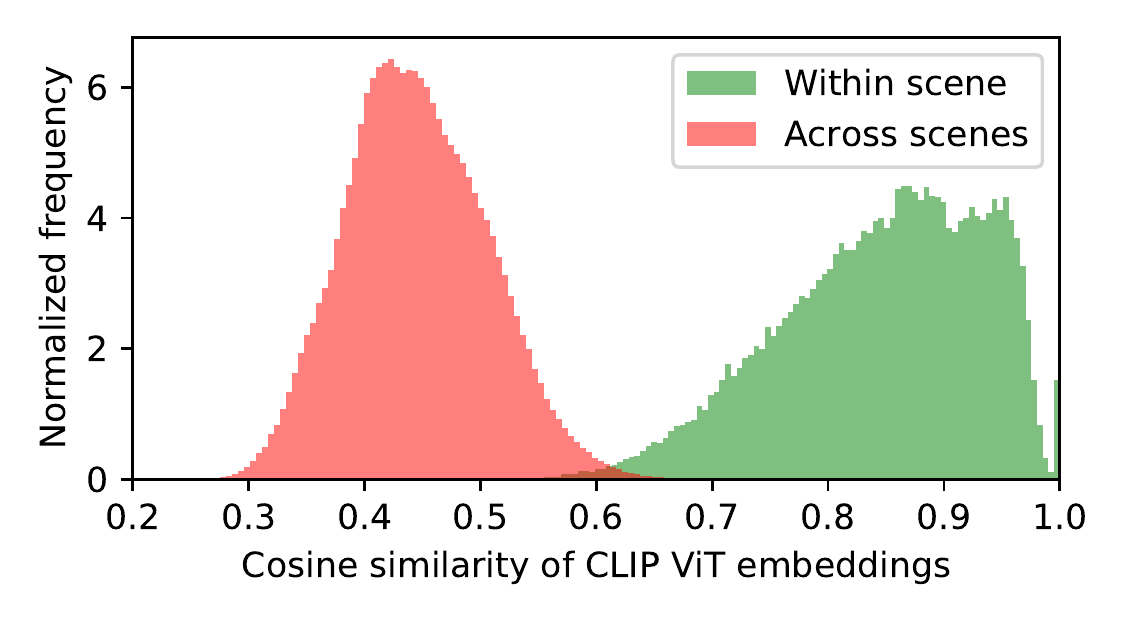}
    \caption{CLIP's Vision Transformer learns low-dimensional image representations through language supervision. We find that these representations transfer well to multi-view 3D settings. We sample pairs of ground-truth views of the same scene and of different scenes from NeRF's Realistic Synthetic object dataset, then compute a histogram of representation cosine similarity. Even though camera poses vary dramatically (views are sampled from the upper hemisphere), views within a scene have similar representations ({\color{green}\textbf{green}}). Across scenes, representations have low similarity ({\color{red}\textbf{red}})}
    \label{fig:representation_similarity_clip_vit}
\end{figure}

In Figure~\ref{fig:representation_similarity_clip_vit}, we measure the pairwise cosine similarity between CLIP representations of views circling an object. We find that pairs of views have highly similar CLIP representations, even for diametrically opposing cameras. This suggests that large, diverse single-view datasets can induce useful representations for multi-view applications.

\subsection{Pose sampling distribution}

We augment the NeRF training loop with $\Lsc$ minimization. Each iteration, we compute $\Lsc$ between a random training image sampled from the observation dataset $I\sim\mathcal{D}$ and rendered image $\hat{I}_\mathbf{p}$ from random pose $\mathbf{p}\sim\pi$. For bounded scenes like NeRF's Realistic Synthetic scenes where we are interested in 360$^\circ$ view synthesis, we define the pose sampling distribution $\pi$ to be a uniform distribution over the upper hemisphere, with radius sampled uniformly in a bounded range. For unbounded forward-facing scenes or scenes where a pose sampling distribution is difficult to define, we interpolate between three randomly sampled known poses $\mathbf{p}_1, \mathbf{p}_2, \mathbf{p}_3 \sim \mathcal{D}$ with pairwise interpolation weights $\alpha_1, \alpha_2 \sim \mathcal{U}(0, 1)$.

\subsection{Improving efficiency and quality}
Volume rendering is computationally intensive. Computing a pixel's color evaluates NeRF's MLP $f_\theta$ at many points along a ray. To improve the efficiency of \ours{} during training, we render images for semantic consistency at low resolution, requiring only 15-20\% of the rays as a full resolution training image. Rays are sampled on a strided grid across the full extent of the image plane, ensuring that objects are mostly visible in each rendering. We found that sampling poses from a continuous distribution was helpful to avoid aliasing artifacts when training at a low resolution.

In experiments, we found that $\Lsc$ converges faster than $\Lmse$ for many scenes. We hypothesize that the semantic consistency loss encourages \ours{} to recover plausible scene geometry early in training, but is less helpful for reconstructing fine-grained details due to the relatively low dimensionality of the ViT representation $\phi(\cdot)$.
We exploit the rapid convergence of $\Lsc$ by only minimizing $\Lsc$ every $k$ iterations. \ours{} is robust to the choice of $k$, but a value between 10 and 16 worked well in our experiments. StyleGAN2~\cite{Karras2019stylegan2} used a similar strategy for efficiency, referring to periodic application of a loss as \textit{lazy regularization}.

As backpropagation through rendering is memory intensive with reverse-mode automatic differentiation, we render images for $\Lsc$ with mixed precision computation and evaluate $\phi(\cdot)$ at half-precision. We delete intermediate MLP activations during rendering and rematerialize them during the backward pass~\cite{chen2016training, MLSYS2020_084b6fbb}. All experiments use a single 16 GB NVIDIA V100 or 11 GB 2080 Ti GPU.

Since $\Lsc$ converges before $\Lmse$, we found it helpful to fine-tune \ours{} with $\Lmse$ alone for 20-70k iterations to refine details.
Alg.~\ref{alg:scarf} details our overall training process.

\section{Experiments}

In experiments, we evaluate the quality of novel views synthesized by \ours{} and baselines for both synthetically rendered objects and real photos of multi-object scenes. (1) We evaluate training \textit{from scratch} on a specific scene with 8 views \S\ref{sec:realistic_synth}. (2) We show that \ours{} improves perceptual quality of view synthesis from \textit{only a single real photo} \S\ref{sec:dtu}. (3) We find that \ours{} can reconstruct regions that are never observed \S\ref{sec:unobserved}, and finally (4) run ablations \S\ref{sec:ablation}.

\textbf{Datasets}~~
The Realistic Synthetic benchmark of~\cite{mildenhall2019llff} includes detailed multi-view renderings of 8 realistic objects with view-dependent light transport effects. We also benchmark on the DTU multi-view stereo (MVS) dataset~\cite{6909453} used by pixelNeRF~\cite{yu2020pixelnerf}. DTU is a challenging dataset that includes sparsely sampled real photos of physical objects.

\textbf{Low-level full reference metrics}~~
Past work evaluates novel view quality with respect to ground-truth from the same pose with Peak Signal-to-Noise Ratio (\textbf{PSNR}) and Structural Similarity Index Measure (\textbf{SSIM})~\cite{sitzmann2019srns}.
PSNR expresses mean-squared error in log space.
However, SSIM often disagrees with human judgements of similarity~\cite{zhang2018perceptual}.

\textbf{Perceptual metrics}~~
Deep CNN activations mirror aspects of human perception.
NeRF measures perceptual image quality using \textbf{LPIPS}~\cite{zhang2018perceptual}, which computes MSE between normalized features from all layers of a pre-trained VGG encoder~\cite{Simonyan15}.
Generative models also measure sample quality with feature space distances. The Fr\'echet Inception Distance (\textbf{FID})~\cite{10.5555/3295222.3295408} computes the Fr\'echet distance between Gaussian estimates of penultimate Inception v3~\cite{44903} features for real and fake images. However, FID is a biased metric at low sample sizes.
We adopt the conceptually similar Kernel Inception Distance (\textbf{KID}), which measures the MMD between Inception features and has an unbiased estimator~\cite{binkowski2018demystifying, obukhov2020torchfidelity}.
All metrics use a different architecture and data than our CLIP ViT encoder.

\subsection{Realistic Synthetic scenes from scratch}
\label{sec:realistic_synth}

\begin{table}
\centering
\caption{Quality metrics for novel view synthesis on subsampled splits of the Realistic Synthetic dataset~\cite{mildenhall2020nerf}. We randomly sample 8 views from the available 100 ground truth training views to evaluate how \ours{} performs with limited observations.}
\label{tab:realistic_synth}
\setlength{\tabcolsep}{1pt}
\begin{tabular}{@{}lccccc@{}}
\toprule
\textbf{Method}         & \textbf{PSNR} $\uparrow$ & \textbf{SSIM} $\uparrow$ & \textbf{LPIPS} $\downarrow$ & \textbf{FID} $\downarrow$ & \textbf{KID} $\downarrow$ \\ \midrule
NeRF    & 14.934 & 0.687  & 0.318 & 228.1 & 0.076 \\
NV &  17.859   & 0.741 & 0.245 & 239.5 & 0.117 \\
Simplified NeRF    & 20.092 & 0.822 & 0.179 & 189.2 & 0.047 \\
\ours{} (ours) & 23.147          & 0.866          & 0.109 & 74.9 & 0.005 \\
\ours{}, $\mathcal{L}_\text{MSE}$ ft  & \textbf{23.591} & \textbf{0.874} & \textbf{0.097} & \textbf{72.0} & \textbf{0.004} \\ \midrule
NeRF, 100 views & \textbf{31.153} & \textbf{0.954} & \textbf{0.046} & \textbf{50.5} & \textbf{0.001} \\ \bottomrule
\end{tabular}
\end{table}

\begin{figure}
    \centering
    \includegraphics[width=\columnwidth]{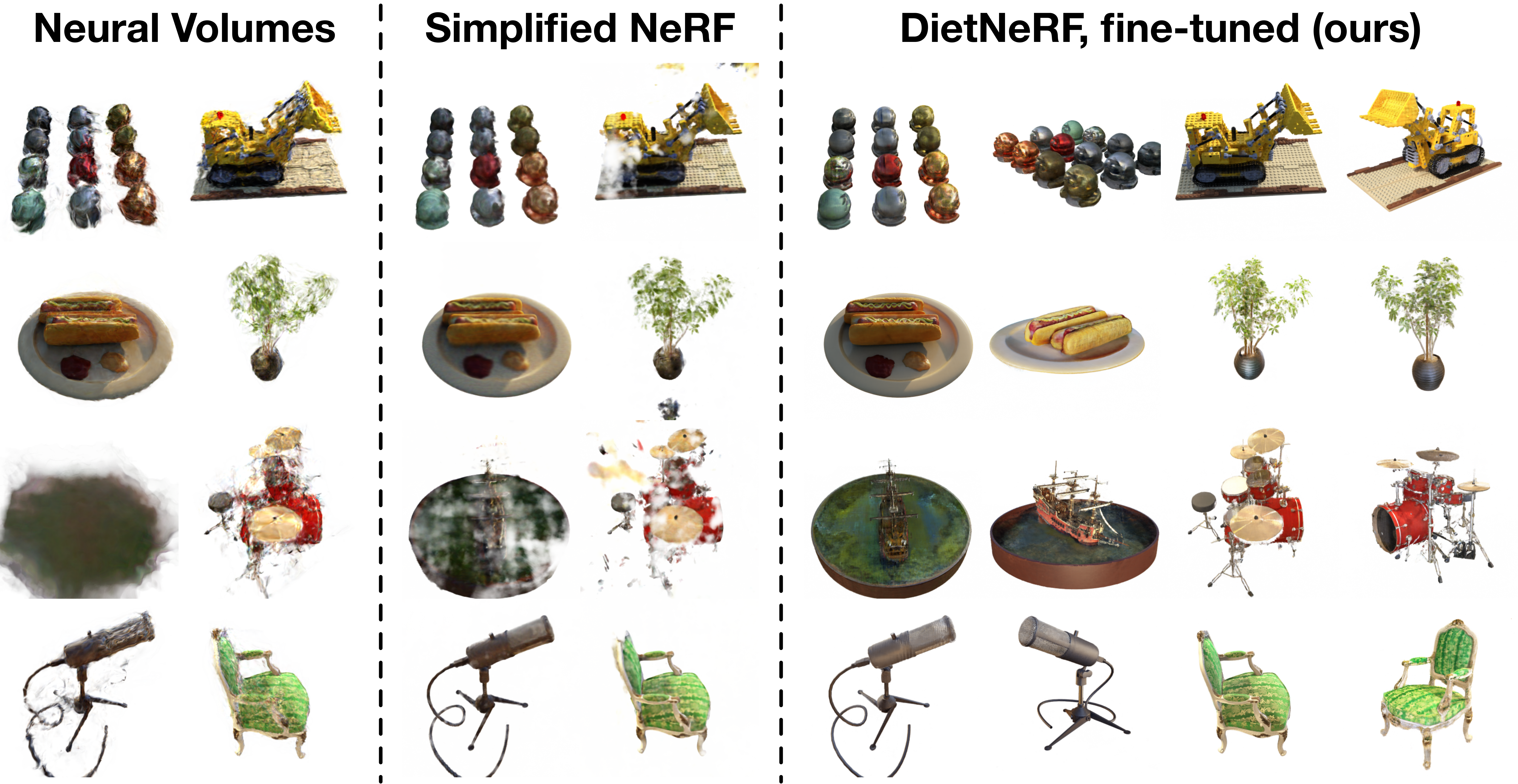}
    \caption{Novel views synthesized from eight observations of scenes in the Realistic Synthetic dataset.}
    \label{fig:realistic_synth_qualitative}
\end{figure}

NeRF's Realistic Synthetic dataset includes 8 detailed synthetic objects with 100 renderings from virtual cameras arranged randomly on a hemisphere pointed inward.
To test few-shot performance, we \textit{randomly} sample a training subset of 8 images from each scene.
Table~\ref{tab:realistic_synth} shows results. The original NeRF model achieves much poorer quantitative quality with 8 images than with the full 100 image dataset. Neural Volumes~\cite{Lombardi:2019} performs better as it tightly constrains the size of the scene's bounding box and explicitly regularizes its scene representation using a penalty on spatial gradients of voxel opacity and a Beta prior on image opacity. This avoids the worst artifacts, but reconstructions are still low-quality. Simplifying NeRF and tuning it for each individual scene also regularizes the representation and helps convergence (+5.1 PSNR over the full NeRF). The best performance is achieved by regularizing with \ours{}'s $\Lsc$ loss. Additionally, fine-tuning with $\Lmse$ even further improves quality, for a total improvement of +8.5 PSNR, -0.2 LPIPS, and -156 FID over NeRF. This shows that semantic consistency is a valuable prior for high-quality few-shot view synthesis. Figure~\ref{fig:realistic_synth_qualitative} visualizes results.

\subsection{Single-view synthesis by fine-tuning}
\label{sec:dtu}

\begin{figure*}[t]
\centering
\includegraphics[width=\linewidth]{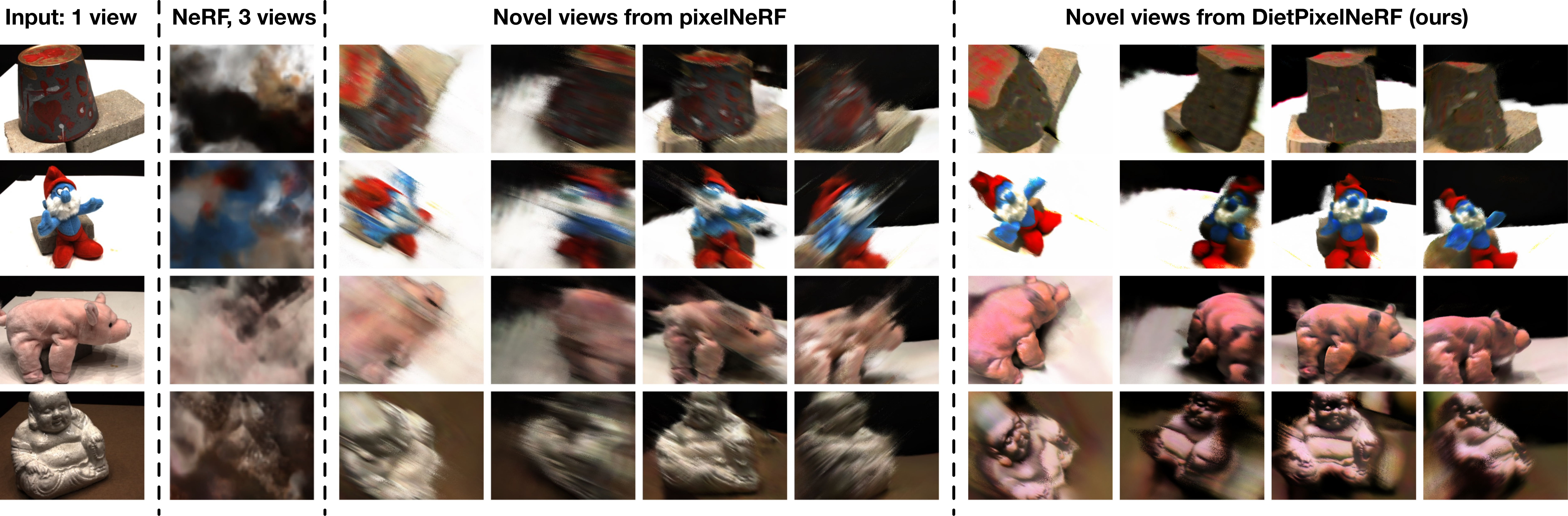}
\caption{\textbf{Novel views synthesized from a \textit{single input image}} from the DTU object dataset. Even with 3 input views, NeRF~\cite{mildenhall2020nerf} fails to learn accurate geometry or textures (reprinted from~\cite{yu2020pixelnerf}). 
While pixelNeRF~\cite{yu2020pixelnerf} has mostly consistent object geometry as the camera pose is varied, renderings are blurry and contain artifacts like inaccurate placement of density along the observed camera's z-axis. In contrast, fine-tuning with \ours{} (\ourspixel{}) learns realistic textures visually consistent with the input image, though some geometric defects are present due to the ambiguous nature of the view synthesis problem.}
\label{fig:dtu}
\end{figure*}

\begin{table}
\caption{\textbf{Single-view novel view synthesis on the DTU dataset}. NeRF and pixelNeRF PSNR, SSIM and LPIPS results are from~\cite{yu2020pixelnerf}. Finetuning pixelNeRF with \ours{}'s semantic consistency loss (\ourspixel{}) improves perceptual quality measured by the deep perceptual LPIPS, FID and KID evaluation metrics, but can degrade PSNR and SSIM which are local pixel-aligned metrics due to geometric defects.}
\label{tab:dtu}
\centering
\setlength{\tabcolsep}{3.4pt}
\begin{tabular}{@{}lccccc@{}}
\toprule
\textbf{Method}         & \textbf{PSNR} & \textbf{SSIM} & \textbf{LPIPS} & \textbf{FID} & \textbf{KID} \\ \midrule
NeRF              & 8.000 & 0.286 & 0.703 & --- & --- \\
pixelNeRF         & 15.550 & 0.537 & 0.535 & 266.1 & 0.166 \\
pixelNeRF, $\mathcal{L}_\text{MSE}$ ft & \textbf{16.048} & \textbf{0.564} & 0.515 & 265.2 & 0.159 \\
\ourspixel{} & 14.242 & 0.481 & \textbf{0.487} & \textbf{190.7} & \textbf{0.066} \\ \bottomrule
\end{tabular}
\end{table}

\begin{figure}
\centering
\includegraphics[width=\columnwidth]{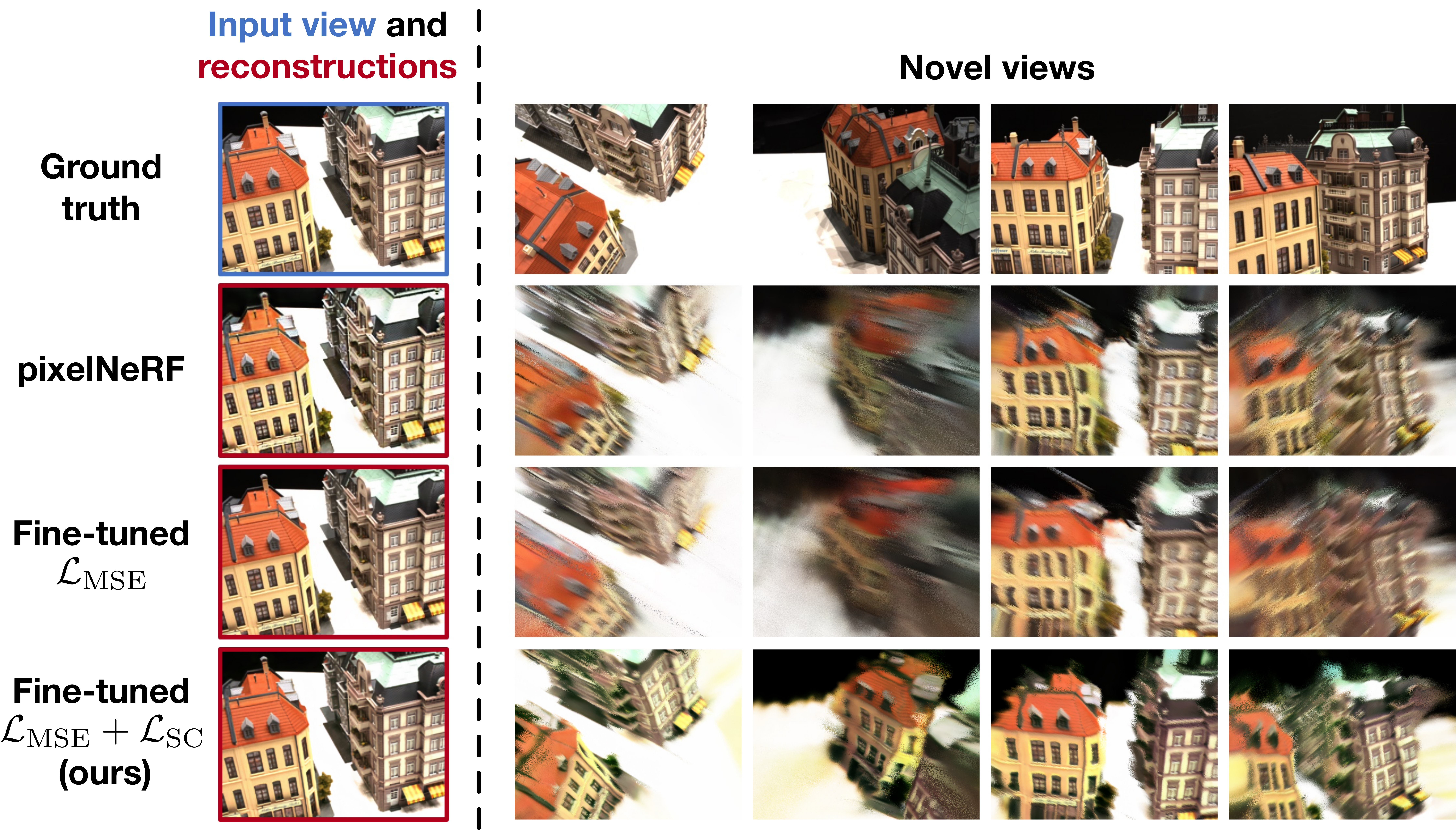}
\caption{\textbf{Semantic consistency improves perceptual quality.} Fine-tuning pixelNeRF with $\mathcal{L}_\text{MSE}$ slightly improves a rendering of the input view, but does not remove most perceptual flaws like blurriness in \textit{novel views}. Fine-tuning with both $\mathcal{L}_\text{MSE}$ and $\mathcal{L}_\text{SC}$ (\ourspixel{}, bottom) improves sharpness of all views.}
\label{fig:dtu_ft_comparison}
\end{figure}

NeRF only uses observations during training, not inference, and uses no auxiliary data. Accurate 3D reconstruction from a single view is not possible purely from $\Lmse$, so NeRF performs poorly in the single-view setting~(Table~\ref{tab:dtu}). 

To perform single- or few-shot view synthesis, pixelNeRF~\cite{yu2020pixelnerf} learns a ResNet-34 encoder and a feature-conditioned neural radiance field on a multi-view dataset of similar scenes. The encoder learns priors that generalize to new single-view scenes. Table~\ref{tab:dtu} shows that pixelNeRF significantly outperforms NeRF given a single photo of a held-out scene. However, novel views are blurry and unrealistic (Figure~\ref{fig:dtu}). We propose to fine-tune pixelNeRF on a single scene using $\Lmse$ alone or using both $\Lmse$ and $\Lsc$. Fine-tuning per-scene with MSE improves local image quality metrics, but only slightly helps perceptual metrics. Figure~\ref{fig:dtu_ft_comparison} shows that pixel-space MSE fine-tuning from one view mostly only improves quality for that view.

We refer to fine-tuning with both losses for a short period as \ourspixel{}. Qualitatively, \ourspixel{} has significantly sharper novel views (Fig.~\ref{fig:dtu}, \ref{fig:dtu_ft_comparison}). \ourspixel{} outperforms baselines on perceptual LPIPS, FID, and KID metrics~(Tab.~\ref{tab:dtu}). For the very challenging single-view setting, ground-truth novel views will contain content that is completely occluded in the input. Because of uncertainty, blurry renderings will outperform sharp but incorrect renderings on average error metrics like MSE and PSNR. Arguably, perceptual quality and sharpness are better metrics than pixel error for graphics applications like photo editing and virtual reality as plausibility is emphasized.

\begin{figure*}[th]
    \centering
    \includegraphics[width=.95\linewidth]{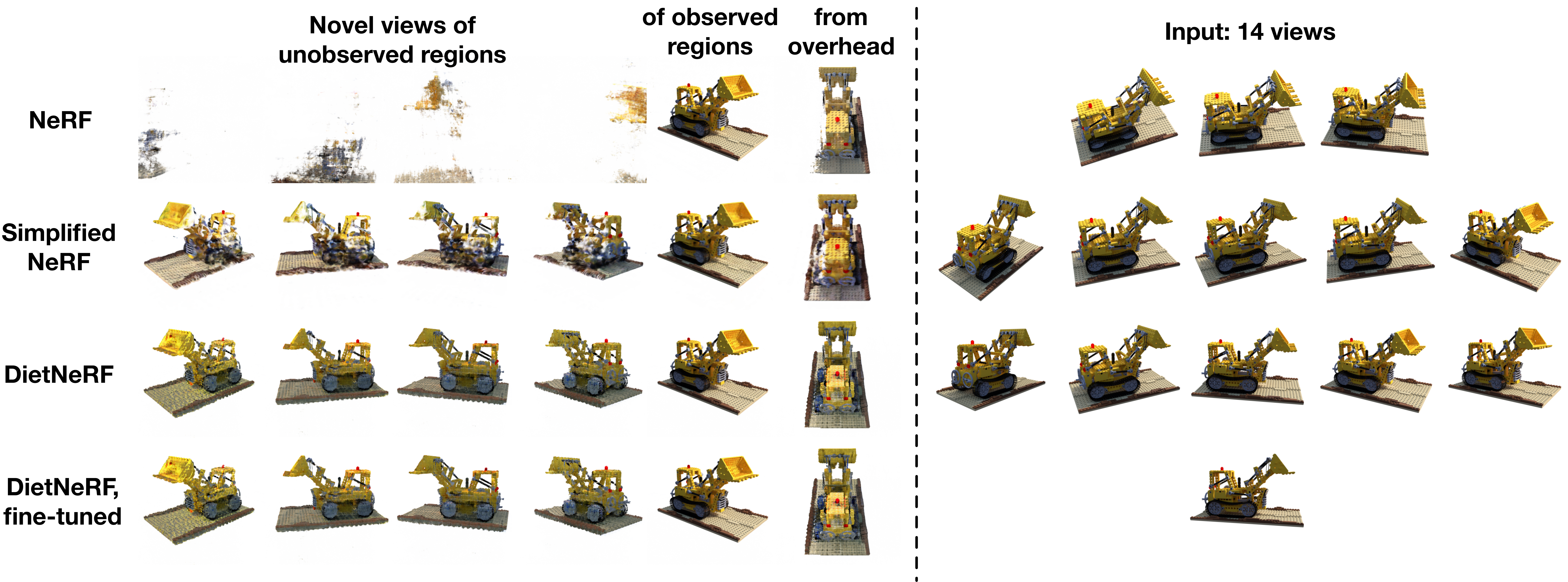}
    \caption{\textbf{Renderings of occluded regions during training.} 14 images of the right half of the Realistic Synthetic lego scene are used to estimate radiance fields. NeRF either learns high-opacity occlusions blocking the left of the object, or fails to generalize properly to the unseen left side. In contrast, \ours{} fills in details for a reconstruction that is mostly consistent with the observed half.}
    \label{fig:unseen}
\end{figure*}

\subsection{Reconstructing unobserved regions}
\label{sec:unobserved}

We evaluate whether \ours{} produces plausible completions when the reconstruction problem is underdetermined. For training, we sample 14 nearby views of the right side of the Realistic Synthetic Lego scene (Fig.~\ref{fig:unseen}, right). Narrow baseline multi-view capture rigs are less costly than 360$^\circ$ captures, and support unbounded scenes. However, narrow-baseline observations suffer from occlusions: the left side of the Lego bulldozer is unobserved. NeRF fails to reconstruct this side of the scene, while our Simplified NeRF learns unrealistic deformations and incorrect colors~(Fig.~\ref{fig:unseen}, left).
Remarkably, \ours{} learns quantitatively (Tab.~\ref{tab:generalization}) and qualitatively more accurate colors in the missing regions, suggesting the value of semantic image priors for sparse reconstruction problems. We exclude FID and KID since a single scene has too few samples for an accurate estimate.

\begin{table}[t]
\centering
\caption{\textbf{Extrapolation metrics.} Novel view synthesis with observations of \textbf{only one side} of the Realistic Synthetic Lego scene.}
\label{tab:generalization}
\setlength{\tabcolsep}{3.2pt}
\begin{tabular}{@{}clccc@{}}
\toprule
\textbf{Views} & \textbf{Method}         & \textbf{PSNR} $\uparrow$ & \textbf{SSIM} $\uparrow$ & \textbf{LPIPS} $\downarrow$ \\ \midrule
14  & NeRF    & 19.662 & 0.799 & 0.202 \\
14  & Simplified NeRF  & 21.553 & 0.818 & 0.160\\
14  & \ours{} (ours)  & 20.753 & 0.810 & 0.157 \\
14  & \ours{} + $\mathcal{L}_\text{MSE}$ ft & \textbf{22.211} & \textbf{0.824} & \textbf{0.143} \\ \midrule
100 & NeRF~\cite{mildenhall2020nerf} & \textbf{31.618} & \textbf{0.965} & \textbf{0.033} \\ \bottomrule
\end{tabular}
\end{table}

\section{Ablations}
\label{sec:ablation}

\textbf{Choosing an image encoder}~~
Table~\ref{tab:supervision_ablation} shows quality metrics with different semantic encoder architectures and pre-training datasets. We evaluate on the Lego scene with 8 views. Large ViT models (ViT L) do not improve results over the base ViT B. Fixing the architecture, CLIP offers a +1.8 PSNR improvement over an ImageNet model, suggesting that data diversity and language supervision is helpful for 3D tasks. Still, both induce useful representations that transfer to view synthesis.

\begin{table}[t]
\centering
\caption{\textbf{Ablating supervision and architectural parameters for the ViT image encoder $\phi(\cdot)$ used to compare image features.} Metrics are measured on the Realistic Synthetic Lego scene.}
\label{tab:supervision_ablation}
\setlength{\tabcolsep}{4.5pt}
\begin{tabular}{@{}lccc@{}}
\toprule
\textbf{Semantic image encoder}     & \textbf{PSNR} $\uparrow$ & \textbf{SSIM} $\uparrow$ & \textbf{LPIPS} $\downarrow$ \\ \midrule
ImageNet ViT L/16, 384$^2$ & 21.501 & 0.809 & 0.167 \\
ImageNet ViT L/32, 384$^2$ & 20.498 & 0.801 & 0.174 \\
ImageNet ViT B/32, 224$^2$ & 22.059 & 0.836 & 0.131 \\
CLIP ViT B/32, 224$^2$     & \textbf{23.896} & \textbf{0.863} & \textbf{0.110} \\\bottomrule
\end{tabular}
\end{table}

\textbf{Varying $\Lmse$ fine-tuning duration}~~
Fine-tuning \ours{} with $\Lmse$ can improve quality by better reconstructing fine-details. In Table~\ref{tab:ft_ablation}, we vary the number of iterations of fine-tuning for the Realistic Synthetic scenes with 8 views. Fine-tuning for up to 50k iterations is helpful, but reduces performance with longer optimization. It is possible that the model starts overfitting to the 8 input views.

\begin{table}[t]
\centering
\caption{\textbf{Varying the number of iterations that \ours{} is fine-tuned with} $\Lmse$ on Realistic Synthetic scenes. All models are initially trained for 200k iterations with $\Lmse$ and $\Lsc$. Further minimizing $\Lmse$ is helpful, but the model can overfit.}
\label{tab:ft_ablation}
\setlength{\tabcolsep}{2.5pt}
\begin{tabular}{@{}lccc@{}}
\toprule
\textbf{Method}     & \textbf{PSNR} $\uparrow$ & \textbf{SSIM} $\uparrow$ & \textbf{LPIPS} $\downarrow$ \\ \midrule
\ours{}, no fine-tuning & 23.147	& 0.866	& 0.109 \\
\ours{}, $\Lmse$ ft 10k	iters  & 23.524	& 0.872	& 0.101 \\
\ours{}, $\Lmse$ ft 50k iters  & \textbf{23.591}	& \textbf{0.874}	& \textbf{0.097} \\
\ours{}, $\Lmse$ ft 100k iters & 23.521	& \textbf{0.874}	& \textbf{0.097} \\
\ours{}, $\Lmse$ ft 200k iters & 23.443	& 0.872	& 0.098 \\\bottomrule
\end{tabular}
\end{table}

\section{Related work}
\textbf{Few-shot radiance fields}~~
Several works condition NeRF on latent codes describing scene geometry or appearance rather than estimating NeRF per scene \cite{schwarz2020graf, grf2020, yu2020pixelnerf}.
An image encoder and radiance field decoder are learned on a multi-view dataset of similar objects or scenes ahead of time. At test time, on a new scene, novel viewpoints are rendered using the decoder conditioned on encodings of a few observed images.
GRAF renders patches of the scene every iteration to supervise the network with a discriminator~\cite{schwarz2020graf}.
Concurrent to our work, IBRNet \cite{wang2021ibrnet} also fine-tunes a latent-conditioned radiance field on a specific scene using NeRF's reconstruction loss, but needed at least 50 views.
Rather than generalizing between scenes through a shared encoder and decoder, \cite{tancik2020learned, Gao-portraitnerf} meta-learn radiance field weights that can be adapted to a specific scene in a few gradient steps. Meta-learning improves performance in the few-view setting. Similarly, a signed distance field can be meta-learned for shape representation problems \cite{sitzmann2020metasdf}.
Much literature studies single-view reconstruction with other, explicit 3D representations. Notable recent examples include voxel~\cite{drcTulsiani17}, mesh~\cite{hu2020worldsheet} and point-cloud~\cite{wiles2020synsin} approaches.

\textbf{Novel view synthesis, image-based rendering}~~
Neural Volumes~\cite{Lombardi:2019} proposes a VAE~\cite{kingma2013auto, pmlr-v32-rezende14} encoder-decoder architecture to predict a volumetric representation of a scene from posed image observations.
NV uses priors as auxiliary objectives like \ours{}, but penalizes opacity based on geometric intuitions rather than RGB image semantics.
TBNs~\cite{olszewski2019tbn} learn an autoencoder with a 3-dimensional latent that can be rotated to render new perspectives for a single-category.
SRNs~\cite{sitzmann2019srns} fit a continuous representation to a scene and also generalize to novel single-category objects if trained on a large multi-view dataset. It can be extended to predict per-point semantic segmentation maps~\cite{kohli2021semantic}.
Local Light Field Fusion~\cite{mildenhall2019llff} estimates and blends multiple MPI representations for each scene.
Free View Synthesis~\cite{10.1007/978-3-030-58529-7_37} uses geometric approaches to improve view synthesis in unbounded in-the-wild scenes.
NeRF++~\cite{kaizhang2020} also improves unbounded scenes using multiple NeRF models and changing NeRF's parameterization.

\textbf{Semantic representation learning}~~
Representation learning with deep supervised and unsupervised approaches has a long history~\cite{bengio2013representation}. Without labels, generative models can learn useful representations for recognition~\cite{chen2020generative}, but self-supervised models like CPC~\cite{oord2019representation,henaff2020dataefficient} tend to be more parameter efficient. Contrastive methods including CLIP learn visual representations by matching similar pairs of items, such as captions and images~\cite{radford2021learning,jia2021scaling}, augmentated variants of an image~\cite{chen2020simple}, or video patches across frames~\cite{jabri2020walk}.

\section{Conclusions}

Our results suggest that single-view 2D representations transfer effectively to challenging, underconstrained 3D reconstruction problems such as volumetric novel view synthesis. While pre-trained image encoder representations have certainly been transferred to 3D vision applications in the past by fine-tuning, the recent emergence of visual models trained on enormous 100M+ image datasets like CLIP have enabled surprisingly effective few-shot transfer. We exploited this transferrable prior knowledge to solve optimization issues as well as to cope with partial observability in the NeRF family of scene representations, offering notable improvements in perceptual quality. In the future, we believe ``diet-friendly'' few-shot transfer will play a greater role in a wide range of 3D applications.

\section*{Acknowledgements}
This material is based upon work supported by the National Science Foundation Graduate Research Fellowship under grant number DGE-1752814 and by Berkeley Deep Drive.
We would like to thank Paras Jain, Aditi Jain, Alexei Efros, Angjoo Kanazawa, Aravind Srinivas, Deepak Pathak and Alex Yu for helpful feedback and discussions.

{\small
\bibliographystyle{ieee_fullname}
\bibliography{main}
}

\clearpage
\appendix

\section{Experimental details}
\paragraph{View selection}
For most few-view Realistic Synthetic experiments, we randomly subsample 8 of the available 100 training renders. Views are not manually selected. However, to compare the ability of NeRF and \ours{} to extrapolate to unseen regions, we manually selected 14 of the 100 views mostly showing the right side of the Lego scene.  For DTU experiments where we fine-tune pixelNeRF~\cite{yu2020pixelnerf}, we use the same source view as~\cite{yu2020pixelnerf}. This viewpoint was manually selected and is shared across all 15 scenes.

\paragraph{Simplified NeRF baseline}

The published version of NeRF~\cite{mildenhall2020nerf} can be unstable to train with 8 views, often converging to a degenerate solution. We found that NeRF is sensitive to MLP parameter initialization, as well as hyperparameters that control the complexity of the learned scene representation. For a fair comparison, we tuned the Simplified NeRF baseline on each Realistic Synthetic scene by modifying hyperparameters until object geometry converged. Table~\ref{tab:simplifications} shows the resulting hyperparameter settings for initial learning rate prior to decay, whether the MLP $f_\theta$ is viewpoint dependent, number of samples per ray queried from the fine and coarse networks, and the maximum frequency sinusoidal encoding of spatial position $(x,y,z)$. The fine and coarse networks are used in~\cite{mildenhall2020nerf} for hierarchical sampling. \xmark~denotes that we do not use the fine network.

\begin{table}[h]
\caption{\textbf{Simplified NeRF training details} by scene in the Realistic Synthetic dataset. We tune the initial learning rate, view dependence,  number of samples from fine and coarse networks for hierarchical sampling, and the maximum frequency of the $(x,y,z)$ spatial positional encoding.}
\label{tab:simplifications}
\centering
\setlength{\tabcolsep}{2.5pt}
\begin{tabular}{@{}lccccc@{}}
\toprule
\textbf{Scene}  & \textbf{LR} & \textbf{View dep.} & \textbf{Fine} & \textbf{Coarse} & \textbf{Max freq.} \\ \midrule
Full NeRF       & $5\times 10^{-4}$ & \cmark  & 128   &  64 & $2^{9}$ \\ \midrule
Lego            & $5\times 10^{-5}$ & \cmark & \xmark & 128 & $2^5$ \\ % 312
Chair           & $5\times 10^{-5}$ & \xmark & \xmark & 128 & $2^5$ \\ % 330
Drums           & $5\times 10^{-5}$ & \xmark & \xmark & 128 & $2^5$ \\ % 331
Ficus           & $5\times 10^{-5}$ & \xmark & \xmark & 128 & $2^5$ \\ % 332
Mic             & $5\times 10^{-5}$ & \xmark & \xmark & 128 & $2^5$ \\ % 334
Ship            & $5\times 10^{-5}$ & \xmark & \xmark & 128 & $2^5$ \\ % 335
Materials       & $1\times 10^{-5}$ & \xmark & \xmark & 128 & $2^5$ \\ % 347
Hotdog          & $1\times 10^{-5}$ & \xmark & \xmark & 128 & $2^3$ \\ % 345
\bottomrule
\end{tabular}
\end{table}

\paragraph{Implementation}
Our implementation is based on a PyTorch port~\cite{YenChen20github_PyTorchNeRF} of NeRF's original Tensorflow code. We re-train and evaluate NeRF using this code.
For memory efficiency, we use 400$\times$400 images of the scenes as in~\cite{YenChen20github_PyTorchNeRF} rather than full-resolution 800$\times$800 images.
NV is trained with full-resolution $800\times800$ views. NV renderings are downsampled with a 2x2 box filter to $400\times400$ to compute metrics.
We train all NeRF, Simplified NeRF and \ours{} models with the Adam optimizer~\cite{DBLP:journals/corr/KingmaB14} for 200k iterations.

\paragraph{Metrics}
Our PSNR, SSIM, and LPIPS metrics use the same implementation as~\cite{yu2020pixelnerf} based on the scikit-image Python package~\cite{scikit-image}. For the DTU dataset, \cite{yu2020pixelnerf} excluded some poses from the validation set as ground truth photographs had excessive shadows due to the physical capture setup. We use the same subset of validation views.

For both Realistic Synthetic and DTU scenes, we also included FID and KID perceptual image quality metrics. While PSNR, SSIM and LPIPS are measured between pairs of pixel-aligned images, FID and KID are measured between two sets of image samples. These metrics compare the \textit{distribution} of image features computed on one set of images to those computed on another set. As distributions are compared rather than individual images, a sufficiently large sample size is needed. For the Realistic Synthetic dataset, we compute the FID and KID between all 3200 ground-truth images (across train, validation and testing splits and across scenes), and 200 rendered test images at the same resolution (25 test views per scene). Aggregating across scenes allows us to have a larger sample size. Due to the setup of the Neural Volumes code, we use additional samples for rendered images for that baseline.
For the DTU dataset, we compute FID and KID between 720 rendered images (48 per scene across 15 validation scenes, excluding the viewpoint of the source image provided to pixelNeRF) and 6076 ground-truth images (49 images including the source viewpoint across 124 training and validation scenes). FID and KID metrics are computed using the \texttt{torch-fidelity} Python package~\cite{obukhov2020torchfidelity}.

\section{Per-scene metrics}

\begin{figure}[t]
    \centering
    \includegraphics[width=\linewidth]{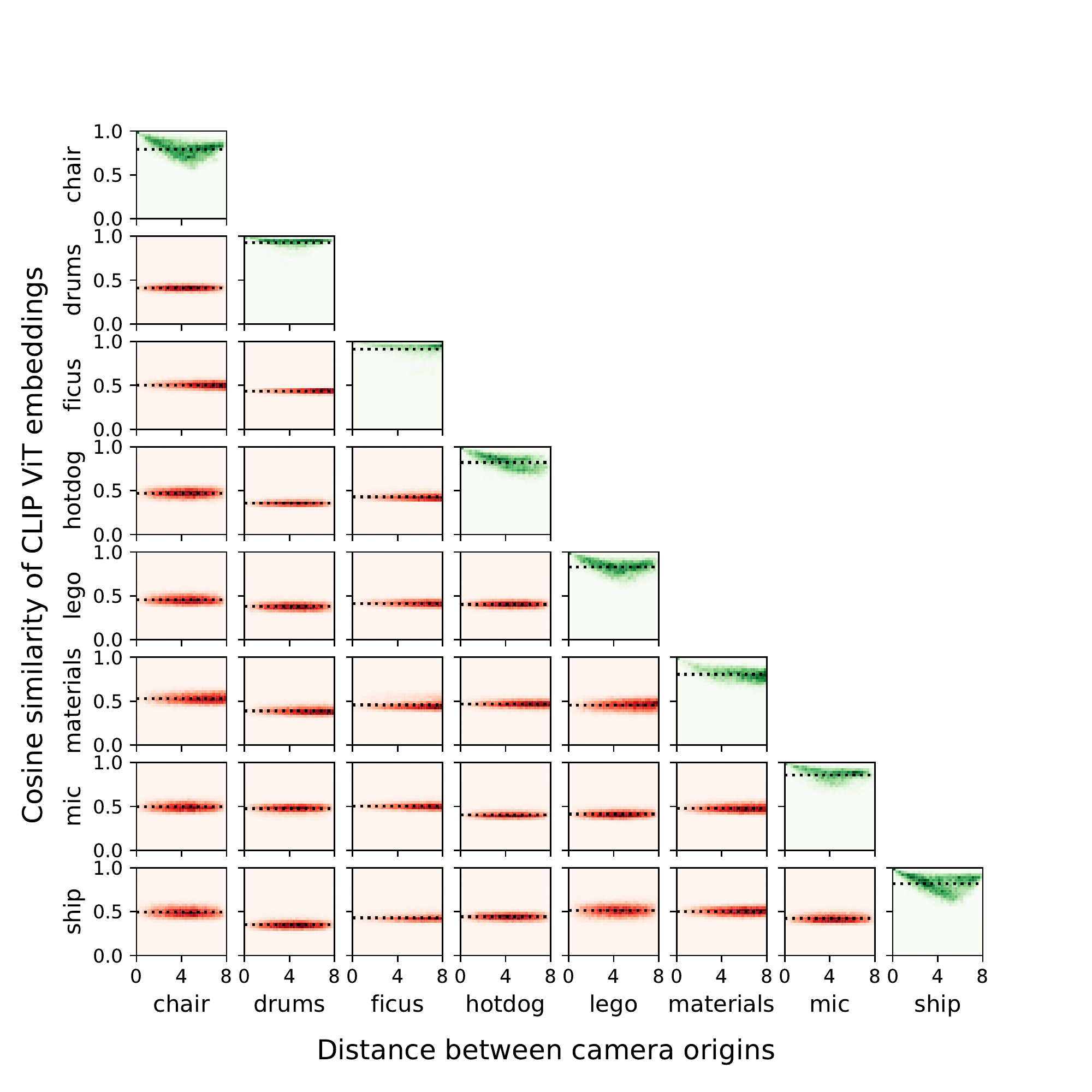}
    \caption{\textbf{CLIP ViT embeddings are more similar between views of the same scene than across different scenes.} We show a 2D histogram for each pair of Realistic Synthetic scenes comparing ViT embedding similarity and the distance between views. The dashed line shows mean cosine similarity, and green histograms have mean similarity is greater than 0.6. On the diagonal, two views from the upper hemisphere of the same scene are sampled. Embeddings of different views of the same scene are generally highly similar. Nearby (distance 0) and diagonally opposing (distance 8) views are most similar. In comparison, when sampling views from different scenes (lower triangle), embeddings are dissimilar.}
    \label{fig:embedding_similarity_grid}
\end{figure}

\begin{table*}
\caption{\textbf{Quality metrics for each scene in the Realistic Synthetic dataset with 8 observed views.}}
\label{tab:per_scene_realistic_synth}
\centering
\setlength{\tabcolsep}{3.65pt}
\begin{tabular}{@{}lcccccccc@{}}
\toprule
\textbf{PSNR} $\uparrow$  & \textbf{Lego} & \textbf{Chair} & \textbf{Drums} & \textbf{Ficus} & \textbf{Mic} & \textbf{Ship} & \textbf{Materials} & \textbf{Hotdog} \\ \midrule
NeRF & 9.726 & 21.049 & 17.472 & 13.728 & 26.287 & 12.929 & 7.837 & 10.446\\
NV~\cite{Lombardi:2019} & 17.652 & 20.515 & 16.271 & 19.448 & 18.323 & 14.457 & 16.846 & 19.361\\
Simplified NeRF & 16.735 & 21.870 & 15.021 & \textbf{21.091} & 24.206 & 17.092 & 20.659 & 24.060\\
DietNeRF (ours) & \underline{23.897} & \underline{24.633} & \textbf{20.034} & 20.744 & \underline{26.321} & \textbf{23.043} & \underline{21.254} & \underline{25.250}\\
DietNeRF, $\Lmse$ ft (ours) & \textbf{24.311} & \textbf{25.595} & \underline{20.029} & \underline{20.940} & \textbf{26.794} & \underline{22.536} & \textbf{21.621} & \textbf{26.626}\\\midrule
NeRF, 100 views & \textbf{31.618} & \textbf{34.073} & \textbf{25.530} & \textbf{29.163} & \textbf{33.197} & \textbf{29.407} & \textbf{29.340} & \textbf{36.899}\\\bottomrule\\
\end{tabular}
\setlength{\tabcolsep}{5pt}
\begin{tabular}{@{}lcccccccc@{}}
\toprule
\textbf{SSIM} $\uparrow$  & \textbf{Lego} & \textbf{Chair} & \textbf{Drums} & \textbf{Ficus} & \textbf{Mic} & \textbf{Ship} & \textbf{Materials} & \textbf{Hotdog} \\ \midrule
NeRF & 0.526 & 0.861 & 0.770 & 0.661 & 0.944 & 0.605 & 0.484 & 0.644\\
NV~\cite{Lombardi:2019} & 0.707 & 0.795 & 0.675 & 0.815 & 0.816 & 0.602 & 0.721 & 0.796\\
Simplified NeRF & 0.775 & 0.859 & 0.727 & \underline{0.872} & 0.930 & 0.694 & 0.823 & 0.894\\
DietNeRF (ours) & \underline{0.863} & \underline{0.898} & \underline{0.843} & \underline{0.872} & \underline{0.944} & \textbf{0.758} & \underline{0.843} & \underline{0.904}\\
DietNeRF, $\Lmse$ ft (ours) & \textbf{0.875} & \textbf{0.912} & \textbf{0.845} & \textbf{0.874} & \textbf{0.950} & \underline{0.757} & \textbf{0.851} & \textbf{0.924}\\\midrule
NeRF, 100 views & \textbf{0.965} & \textbf{0.978} & \textbf{0.929} & \textbf{0.966} & \textbf{0.979} & \textbf{0.875} & \textbf{0.958} & \textbf{0.981}\\\bottomrule\\
\end{tabular}
\begin{tabular}{@{}lcccccccc@{}}
\toprule
\textbf{LPIPS} $\downarrow$  & \textbf{Lego} & \textbf{Chair} & \textbf{Drums} & \textbf{Ficus} & \textbf{Mic} & \textbf{Ship} & \textbf{Materials} & \textbf{Hotdog} \\ \midrule
NeRF & 0.467 & 0.163 & 0.231 & 0.354 & 0.067 & 0.375 & 0.467 & 0.422\\
NV~\cite{Lombardi:2019} & 0.253 & 0.175 & 0.299 & 0.156 & 0.193 & 0.456 & 0.223 & 0.203\\
Simplified NeRF & 0.218 & 0.152 & 0.280 & 0.132 & 0.080 & 0.283 & 0.151 & 0.139\\
DietNeRF (ours) & \underline{0.110} & \underline{0.092} & \textbf{0.117} & \underline{0.097} & \underline{0.053} & \underline{0.204} & \underline{0.102} & \underline{0.097}\\
DietNeRF, $\Lmse$ ft (ours) & \textbf{0.096} & \textbf{0.077} & \textbf{0.117} & \textbf{0.094} & \textbf{0.043} & \textbf{0.193} & \textbf{0.095} & \textbf{0.067}\\\midrule
NeRF, 100 views & \textbf{0.033} & \textbf{0.025} & \textbf{0.064} & \textbf{0.035} & \textbf{0.023} & \textbf{0.125} & \textbf{0.037} & \textbf{0.025}\\\bottomrule
\end{tabular}
\end{table*}

\paragraph{Embedding similarity}
In Figure~\ref{fig:embedding_similarity_grid}, we compare the cosine similarity of two views with the distance between their camera origins for each pair of scenes in the Realistic Synthetic dataset. When sampling both views from the same scene, views have high cosine similarity (diagonal). For 6 of the 8 scenes, there is some dependence on the relative poses of the camera views, though similarity is high across all camera distances. For views sampled from different scenes, similarity is low (cosine similarity around 0.5).

\paragraph{Quality metrics}
Table~\ref{tab:per_scene_realistic_synth} shows PSNR, SSIM and LPIPS metrics on a per-scene basis for the Realistic Synthetic dataset. FID and KID metrics are excluded as they need a larger sample size. We bold the best method on each scene, and underline the second-best method. Across all scenes in the few-shot setting, \ours{} or \ours{} fine-tuned for 50k iterations with $\Lmse$ performs best or second-best.

\section{Qualitative results and ground-truth}
\begin{figure}[t]
    \centering
    \includegraphics[width=\linewidth]{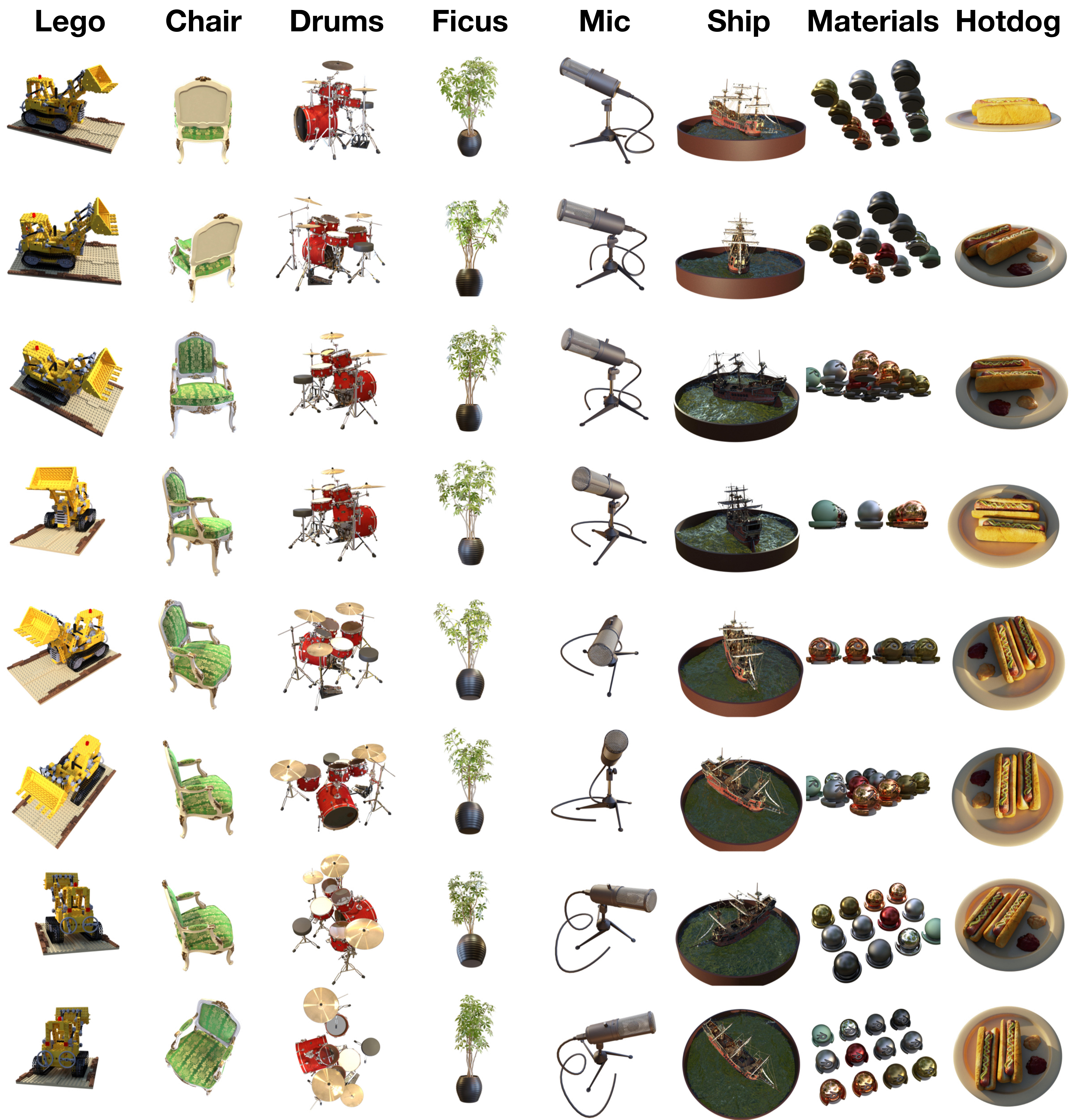}
    \caption{\textbf{Training views used for Realistic Synthetic scenes.} These views are randomly sampled from the available 100 views. This is a challenging setting for view synthesis and 3D reconstruction applications as objects are not uniformly observed. Some views are mostly redundant, like the top two Lego views. Other regions are sparsely observed, such as a single side view of Hotdog.}
    \label{fig:realistic_synth_training_views}
\end{figure}

In this section, we provide additional qualitative results. Figure~\ref{fig:realistic_synth_training_views} shows the ground-truth training views used for 8-shot Realistic Synthetic experiments. These views are sampled at random from the training set of~\cite{mildenhall2020nerf}. Random sampling models challenges with real-world data capture such as uneven view sampling. It may be possible to improve results if views are carefully selected.

\begin{figure*}[p]
    \centering
    \includegraphics[width=0.9\linewidth]{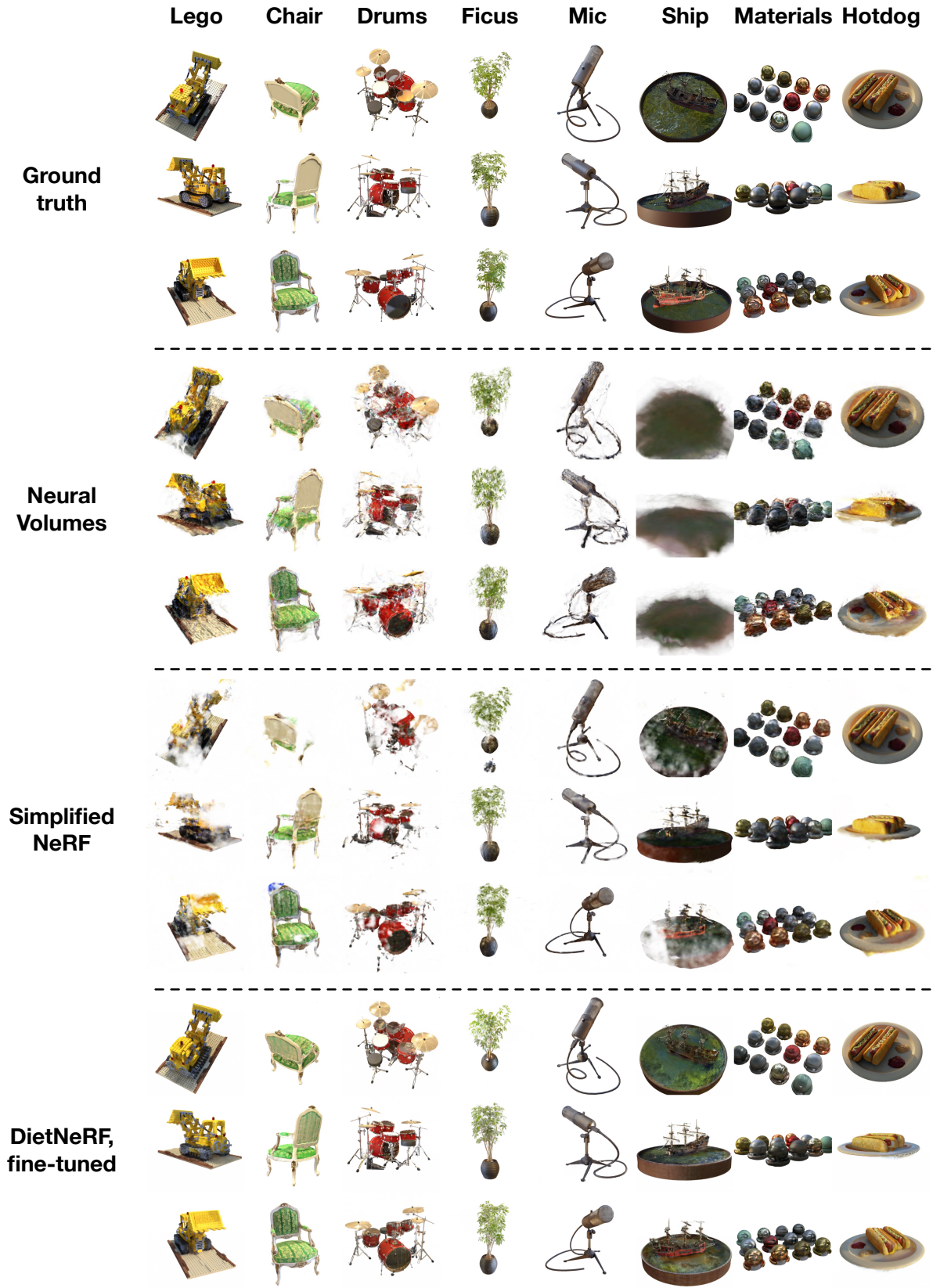}
    \caption{\textbf{Additional renderings of Realistic Synthetic scenes.}}
    \label{fig:realistic_synth_qual_extra}
\end{figure*}

In Figure~\ref{fig:realistic_synth_qual_extra}, we provide additional renderings of Realistic Synthetic scenes from testing poses for baseline methods and \ours{}. Neural Volumes generally converges to recover coarse object geometry, but has wispy artifacts and distortions. On the Ship scene, Neural Volumes only recovers very low-frequency detail. Simplified NeRF suffers from occluders that are not visible from the 8 training poses. \ours{} has the highest quality reconstructions without these distortions or occluders, but does miss some high-frequency detail. An interesting artifact is the leakage of green coloration to the back of the chair.

\begin{figure*}[p]
    \centering
    \includegraphics[width=\linewidth]{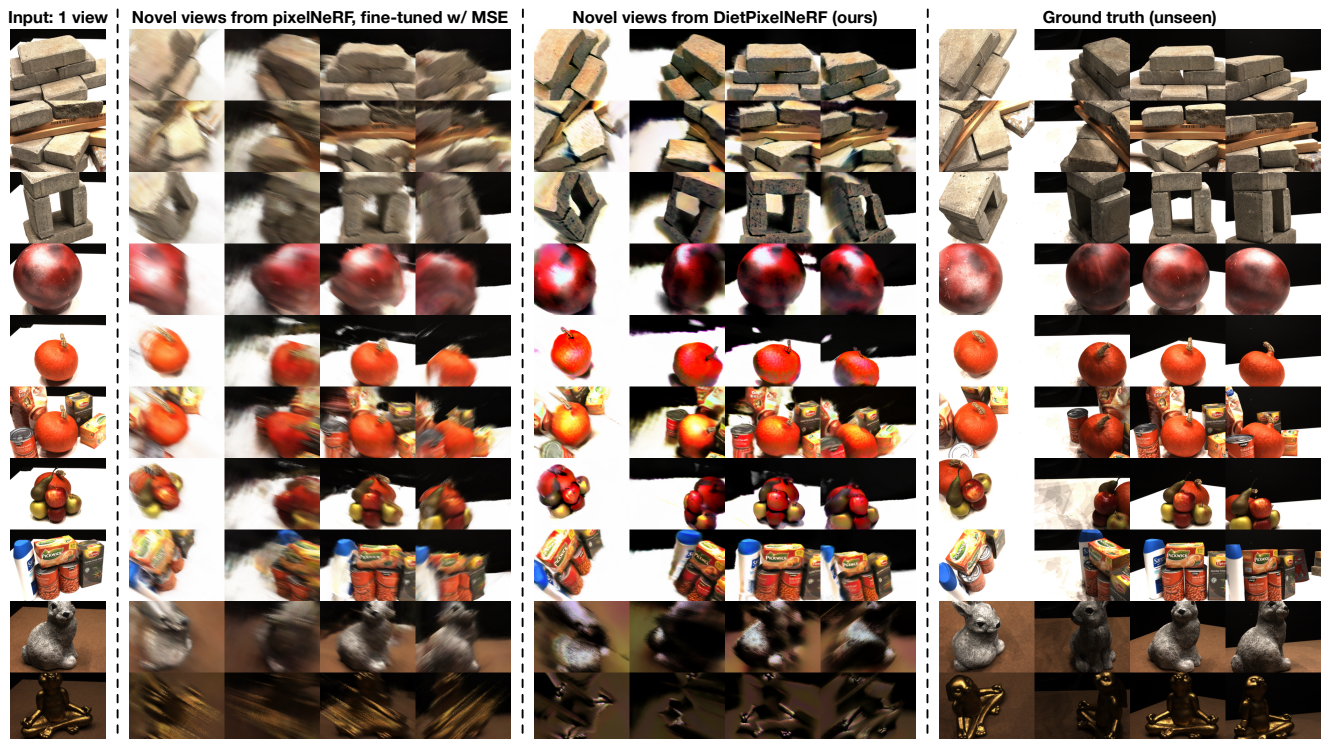}
    \caption{\textbf{One-shot novel view synthesis:} Additional renderings of DTU scenes generated from a single observed view (left). Ground truth views are shows for reference, but are not provided to the model. pixelNeRF and \ourspixel{} are pre-trained on the same dataset of other scenes, then fine-tuned on the single input view for 20k iterations with $\Lmse$ alone (pixelNeRF) or $\Lmse + \Lsc$ (\ourspixel{}).}
    \label{fig:dtu_qual_extra}
\end{figure*}

Finally, in Figure~\ref{fig:dtu_qual_extra}, we show renderings from pixelNeRF and \ourspixel{} on all DTU dataset validation scenes not included in the main paper. Starting from the same checkpoint, pixelNeRF is fine-tuned using $\Lmse$ for 20k iterations, whereas \ourspixel{} is fine-tuned using $\Lmse + \Lsc$ for 20k iterations. \ourspixel{} has sharper renderings. On scenes with rectangular objects like bricks and boxes, \ourspixel{} performs especially well. However, the method struggles to preserve accurate geometry in some cases. Note that the problem is under-determined as only a single view is observed per scene.

\section{Adversarial approaches}

While NeRF is only supervised from observed poses, conceptually, a GAN~\cite{NIPS2014_5ca3e9b1} uses a discriminator to compute a realism loss between real and generated images that need not align pixel-wise. Patch GAN discriminators were introduced for image translation problems~\cite{pix2pix2017, CycleGAN2017} and can be useful for high-resolution image generation~\cite{esser2020taming}. SinGAN~\cite{rottshaham2019singan} trains multiscale patch discriminators on a single image, comparable to our single-scene few-view setting. 
In early experiments, we trained patch-wise discriminators per-scene to supervise $f_\theta$ from novel poses in addition to $\Lsc$. However, an auxiliary adversarial loss led to artifacts on Realistic Synthetic scenes, both in isolation and in combination with our semantic consistency loss.

\end{document}